\documentclass{article}
\usepackage{ijcai24}

\usepackage{times}
\usepackage{soul}
\usepackage{url}
\usepackage[hidelinks]{hyperref}
\usepackage[utf8]{inputenc}
\usepackage[small]{caption}
\usepackage{graphicx}
\usepackage{amsmath}
\usepackage{amsthm}
\usepackage{booktabs}
\usepackage{algorithm}
\usepackage{algorithmic}
\usepackage[switch]{lineno}
\usepackage{xcolor}

\title{FMEA Builder: Expert Guided Text Generation for Equipment Maintenance
\footnote{
Please cite this as: 
Karol Lynch, Fabio Lorenzi, John Sheehan, Duygu Kabakci-Zorlu, Bradley Eck. 
FMEA Builder: Expert Guided Text Generation for Equipment Maintenance.
In \emph{AI for Critical Infrastructure Workshop @ IJCAI-24}, 2024/8/3.}
}

\author{
Karol Lynch\and
Fabio Lorenzi\and
John Sheehan\and
Duygu Kabakci-Zorlu\And
Bradley Eck
\affiliations
IBM Research Europe, Dublin, Ireland\\
\emails
karol\_lynch@ie.ibm.com, fabio.lorenzi1@ibm.com, john.d.sheehan@ie.ibm.com, duygu.kabakci.zorlu@ibm.com, bradley.eck@ie.ibm.com
}

\begin{document}

\maketitle

\begin{abstract}
Foundation models show great promise for generative tasks in many domains. Here we discuss the use of foundation models to generate structured
documents related to critical assets.
A Failure Mode and Effects Analysis (FMEA) captures the composition of an asset or piece of equipment, the ways it may fail and the consequences thereof.
Our system uses large language models to enable fast and expert supervised generation of new FMEA documents. 
Empirical analysis shows that foundation models can correctly generate over half of an FMEA's key content. 
Results from polling audiences of reliability professionals show a positive outlook on using generative AI to create these documents for critical assets. 
\end{abstract}

\section{Introduction}
We propose an AI system for the generation of structured documents related
to industrial equipment with particular focus on Failure Mode and Effects
Analysis (FMEA). 
FMEAs are a longstanding tool of reliability engineering for understanding equipment failure points and optimal maintenance strategies~\cite{srt2004,sharma:JARASS-2018}. These documents capture reasons why equipment, assets and infrastructures fail and outline maintenance options for such failures 
to achieve the desired level of reliability.

Although FMEAs content can differ by sector, our approach considers a document with the following sections: 
\begin{itemize}
\item The boundary gives a functional description along with the main components. 
\item Failure locations are points on the equipment where a failure might occur. 
\item Degradation mechanisms describe the physical process or mechanism that can lead to a failure. 
\item Degradation influences describe the underlying causes of a degradation.
\item Preventative maintenance tasks can be carried out to prevent
failures; and,
\item Job plans collect such tasks into a schedule.
\end{itemize}
These parts of the document show a nested behaviour wherein a typical
piece of equipment has multiple failure locations with each
being linked with one or more degradation mechanisms and so on. Preventative activities depend on the failure location, mechanism, and influence. Job
plans schedule preventative activities according to operating
conditions, while usually grouping related preventative steps 
together.

Such an approach is regarded as being effective to managing critical infrastructure
in many sectors including Energy and Utilities, Water and Wastewater management, and Oil and Gas ~\cite{10062421} where the effect of unforseen, unplanned or disastrous failures without solid recovery strategies has severe impacts on the system, business and potentially society as a whole.

Creating an FMEA requires a group of highly trained experts focusing on a single study. Such resources might be too expensive or unavailable to some organisations involved in critical infrastructure management.  

Generating FMEAs is challenging because the sequential relationship of the document's sections can propagate errors and because FMEAs contain domain specific knowledge about the equipment and how it is used. 
In addition, the same words can refer to different equipment components, with the correct interpretation depending on the usage.  
For example the ``casing" of a pump is different than that of a window. This behavior makes it difficult to create a navigable catalog of components from which to build FMEAs. However, the attention mechanism \cite{Vaswani-2017} used in today's language models can interpret the meaning of words based on their context. 

In this discussion we explore how large language models (LLMs)~\cite{bubeck:arxiv-2023} can assist in the creation of FMEAs.  
Our system for generating FMEAs draws on recent techniques for using LLMs
and contributes a case study for using LLMs on domain-specific problems.
Key techniques informing our work include answer consistency~\cite{wang:arxiv-2023}, in-context learning~\cite{Brown:neurips-2020}, and dynamic relevant example selection~\cite{liu:ACL-2022,nori:arxiv-2023}.
Recent studies applying LLMs to particular domains include the work of Nori et al.~\cite{nori:arxiv-2023} for medical tasks and Balaguer et al.~\cite{balaguer:arxiv-2024}  for agriculture.
Those studies respectively showed that innovative prompting could achieve state of the art performance and that retrieval augmented generation and fine-tuning both had advantages for domain-specific problems.

With this landscape, our main contributions are
 experiments comparing the performance of several LLMs on generating parts of an FMEA. We also share feedback on the work in progress  
design for an interactive system that enables convenient collaboration between LLMs and human experts for FMEA creation. 

In the remainder of the paper, we describe our approach to generating FMEAs and share results comparing different LLMs for the task. We also share survey feedback from target users of our current prototype. 
We conclude with a summary of the future directions planned for exploration.

\section{Solution Approach}
Our system for creating new FMEAs decomposes the generation problem according to the structure of the documents we generate. This decomposition enables our target users, subject matter experts, to inject their knowledge and supervise the generation process.
We use a library of existing documents to furnish the model with relevant information, and generate structured outputs for consumption by our graphical interface.
Taken together, these methods should enable experts to create new FMEAs of high quality in a short time.

\begin{figure}
\includegraphics[width=\columnwidth]{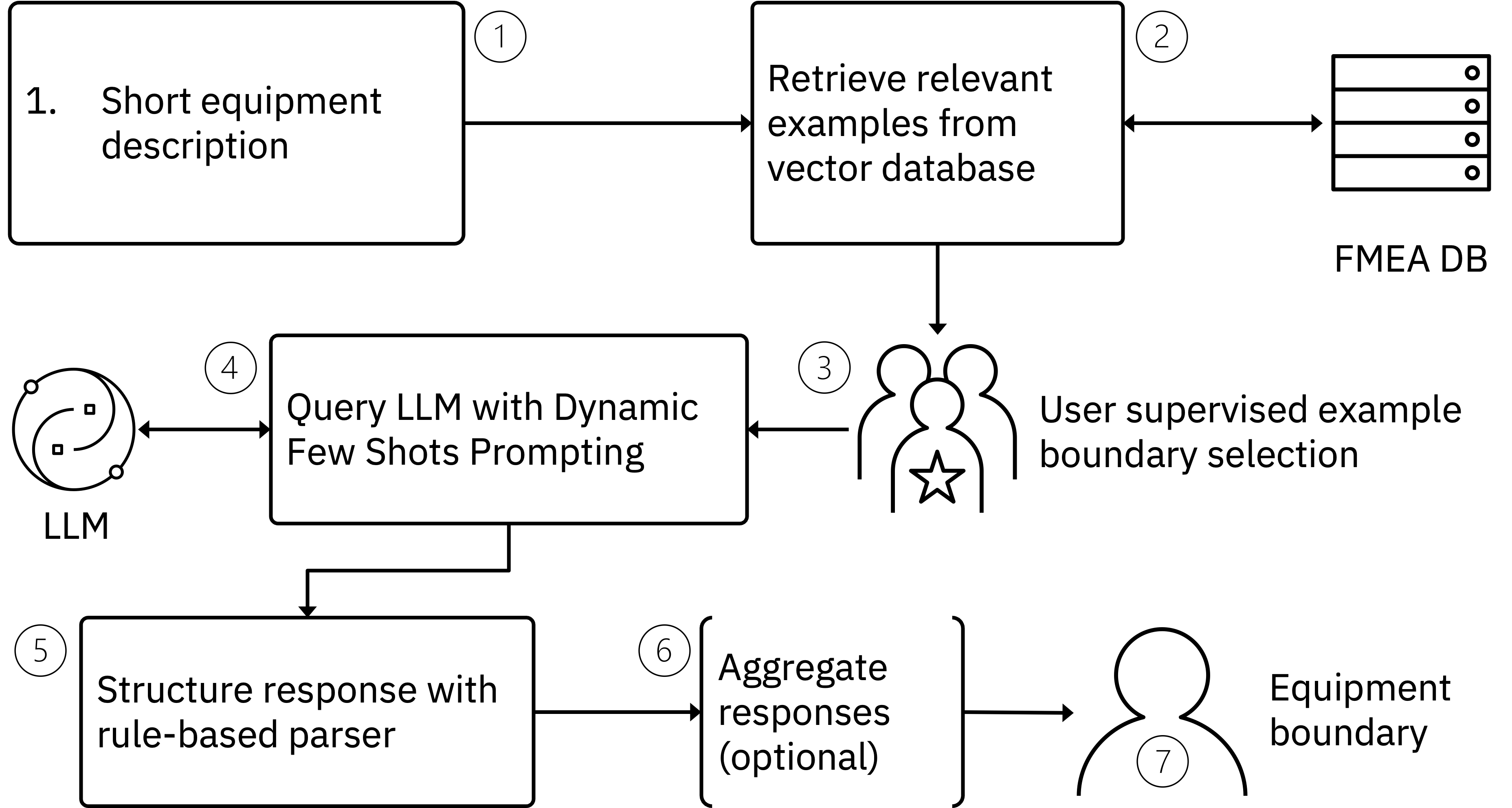}
\caption{Flow of a single step (boundary generation) in our system: The step's input (1) is used to select candidate examples from the database (2). Examples confirmed by the user (3) appear in the prompt (4). The system parses the LLM's response (5) and optionally aggregates responses from multiple prompt, model variations (6) for presentation to the user (7).}
\label{fig:solution-flow}
\end{figure}

The workflow for the first step, generating the equipment boundary from a short description, is a representative example for generating part of an FMEA (Fig. \ref{fig:solution-flow}). 
The prompt construction and response parsing steps are further described  below. 

\paragraph{Dynamic Few Shot Prompting (DFSP)} We rank and retrieve relevant examples, or shots, by cosine similarity of text embeddings. 
This approach, termed 
Dynamic Few Shot Prompting, adds user supervision to the example selection methods of 
~\cite{liu:ACL-2022,nori:arxiv-2023}. DFSP combines three sources of knowledge: examples already in the database, an expert's knowledge, and the knowledge in the LLM. 
Running the prompts without injected examples provides a \emph{zero-shot} method that draws only on the LLM. 

\paragraph{Structured responses} Parsing the LLM's response from pure text into structured responses enables presentation of FMEA components  directly to the user for subsequent supervision. For example the user can confirm, reject or supplement the generated list of failure locations. 
Structured responses also simplify the resolution of repeated entities, a common problem in text generation~\cite{holtzman:arxiv-2020}. 
Our system generates structured responses by lightly formatting the injected examples with simple delimiters. A rule-based parser operates over the delimiters to produce a response in javascript object notation.

\section{Evaluation}\label{sec:eval}

For the work in progress reported here we focus on the first two parts of an FMEA document:  
the \emph{equipment boundary} and, the \emph{failure locations}. 
Experiments used our propriety database of 714 FMEAs developed by subject matter experts and a
range of state of the art LLMs: llama-2-70b-chat (70B)~\cite{touvron:arxiv-2023}, flan-ul2 (20B)~\cite{tay:arxiv-2023}, and quantizated versions of Mixtral-8x7B (8x7B)~\cite{jiang:arxiv-2024}, denoted by Mixtral-Q. 

For evaluation purposes we divided the database into train (n=571;
80\%), validation (n=71; 10\%) and test (n=72; 10\%) splits. Examples for our DFSP method are drawn from the training split. 
We compare our DFSP method with prompts using a randomly selected example (\emph{random-shot}) and no example (\emph{zero shot}). The random shot uses a random example from the database to provide  a syntactic rather than semantic hint to the model.

Although it is not possible to share data for reproducing our experiments we suggest that results reported here can provide valuable insights for the domain.

\subsection{Generating Equipment Boundaries}
An equipment boundary of an industrial asset describes the asset and its constituent components.
The input is a few-word description of the equipment and the output is the boundary itself.

We evaluated the performance of several models for generating equipment boundaries in terms of the ROUGE-1~\cite{lin-2004-rouge} score for the unstructured response, and recall and precision for the structured list of components. ROUGE-1 is a recall oriented similarity score for comparing candidate and reference texts with values ranging from zero to one.

Results (Table \ref{tab:boundary}) for individual models showed a clear pattern of increasing quality as we move from zero-shot to random-shot and DFSP, with flan-ul2 the best performing individual model. Results for DFSP show a strong uplift for ROUGE-1 and for component lists.
Although the described methods automatically generate much of the required information, some information, in particular on the component lists, is missed. This result
emphasizes the important role of reliability engineers to supervise the
generated descriptions.

\begin{table}
\centering
\caption{Performance for generating equipment boundaries on test split (n=72).}
\begin{tabular}{llccc}
\hline
Model    & Method    & ROUGE-1 & Recall   & Prec  \\
\hline
flan-ul2     &  zero-shot   & 0.133  & 0.080   & 0.135  \\
flan-ul2     &  random-shot   & 0.274 & 0.113   & 0.147   \\
flan-ul2     &  DFSP   & \textbf{0.787}  & \textbf{0.573}   & \textbf{0.546} \\
\hline
llama2       &  zero-shot   &  0.299 & 0.075    & 0.082    \\
llama2     &  random-shot   & 0.357  & 0.107   & 0.076  \\
llama2       &  DFSP    & 0.685  & 0.483   & 0.415  \\
\hline
mixtral-Q  &  zero-shot  & 0.238  & 0.050   & 0.044   \\
mixtral-Q    &  random-shot   & 0.349  & 0.092  & 0.056  \\
mixtral-Q     &  DFSP  & 0.573  & 0.342   & 0.327  \\
\hline
\end{tabular}
\label{tab:boundary}
\end{table}

\subsection{Generating Failure Locations}
The set of failure locations for an industrial asset is a key part of an FMEA indicating the components of an asset likely to fail. The input to this step is the equipment boundary. 
This evaluation uses boundaries from the database to generate failure locations.
Results showed a similar pattern as equipment boundaries;
quality increases from zero-shot to random-shot to DFSP (Table~\ref{tab:locn}).
Results for individual models were more mixed with llama2 showing the highest recall and F1 while mixtral-Q had the highest precision.

\begin{table}
\centering
\caption{Performance for generating failure locations on test split (n=72).}
\begin{tabular}{ll ccc}
\hline
Model      & Method     & Recall &  Prec  & F1\\
\hline
flan-ul2 & zero-shot             & 0.031    & 0.139 &  0.051 \\
flan-ul2 & random-shot           & 0.176    & 0.351 &  0.234 \\
flan-ul2 & DFSP	              & 0.454    & 0.585 &  0.511 \\
\hline
llama2 & zero-shot             & 0.229   & 0.274 &  0.250\\
llama2 & random-shot           & 0.243    & 0.253 & 0.248\\
llama2 & DFSP	              & \textbf{0.559}    & 0.597 & \textbf{0.577} \\
\hline
mixtral-Q & zero-shot             & 0.040    & 0.167 & 0.065 \\
mixtral-Q & random-shot           & 0.121    & 0.271 & 0.168 \\
mixtral-Q & DFSP	              & 0.482    & \textbf{0.612} & 0.539 \\
\hline
\label{tab:locn}
\end{tabular}
\end{table}

\subsection{User Feedback}

\begin{figure}
\includegraphics[width=\columnwidth]{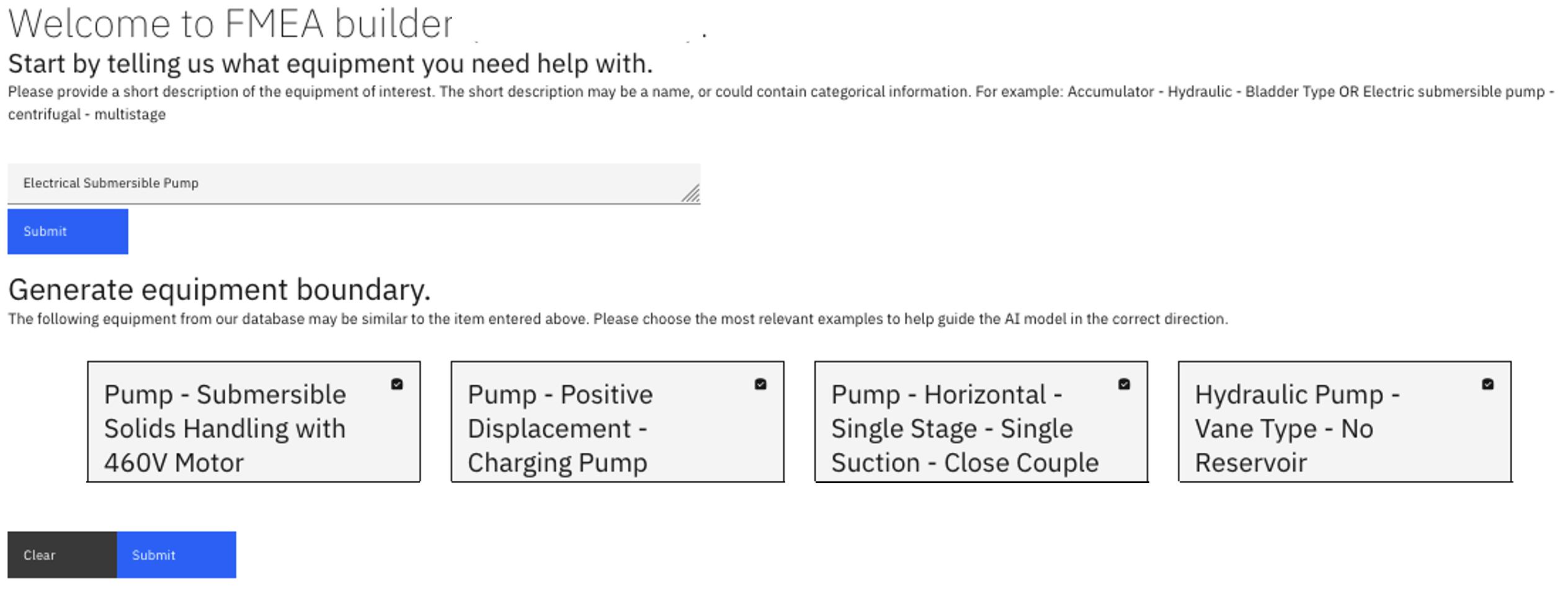}
\caption{Graphical interface for generating equipment boundaries: The user enters a short description of the equipment as free text. Examples of similar equipment from our database are presented for consideration. Equipment cards that are ticked serve as examples in the prompt to generate an equipment boundary.}
\label{fig:boundary}
\end{figure}

Our system enables user interaction with structured model responses through a tailored graphical interface. As an example, our interface for the first step in the pipeline appears in Figure \ref{fig:boundary}. 

We showed the user interface for this work in progress system to two audiences of people responsible for the maintenance and reliability of critical infrastructure across industries. 
For these professionals, creation and application of FMEAs form a crucial part of their daily work.  Following the demonstrations, we polled the audience for feedback on two key questions.

\begin{description}
\item
Question 1: \emph{How likely would you be to use a tool like the demo during FMEA creation if it was available?} Answer options: Extremely likely; Likely; Undecided; Unlikely; Extremely unlikely.
\item
Question 2: \emph{How much configurability would you like to have when using the tool during FMEA creation?} Answer options: Build FMEAs fully automated by AI; Build FMEAs mostly automated by AI; Build FMEAs acting as my helpful but supervised assistant; Build FMEAs mostly manually; Build FMEAs fully manually.
\end{description}

Table~\ref{tab:user} reports responses favorable to using the tool and to having the support of AI in general. In both audiences, responses were positive. We interpret higher scores for AI in general than for our tool in particular as an opportunity to improve the design of the interface and to familiarize users with the capabilities and limitations of AI.

\begin{table}
\centering
\caption{User Survey Feedback}
\begin{tabular}{cccc}
\hline
Audience & Size & Positive [Q1] & Positive [Q2] \\
\hline
A        & 27   &  82\%       & 96\% \\
B        & 55   &     -        & 98\%  \\
\hline
\end{tabular}
\label{tab:user}
\end{table}

\section{Outlook}

In this work we have shared a view of current work in progress for
generating documents related to critical equipment. 
The method of dynamically retrieving examples from a database for including 
in a prompt shows
the ability to correctly generate over half of the content needed for an FMEA.  Although this level of performance 
is considered helpful according to our surveys, we intend to explore 
further improvements. 
In particular, ensemble methods based on fuzzy voting provide a promising tool for combining results between models and shot orderings. 
We expect ensembles to improve the recall of structured responses at a hopefully small cost in precision. 

So far our experiments have focused on the test split from the database but there are many equipment types not covered by the database where FMEA generation remains of interest. In these cases, knowledge is often available 
in the less structured form of manuals and process documents. With pre-processing / chunking this information can also be used to generate parts of an FMEA. 
Ultimately we foresee the use of ensemble methods to combine results between LLM responses informed by examples from the database as well as user-provided manuals and documents. 

Finally further feedback sessions remain to be conducted during the development of this project to evaluate the perceived quality of the generated documents as opposed to the internal algorithmic evaluation.

\bibliographystyle{named}
\bibliography{ours}

\end{document}